\documentclass{article}

\usepackage[nonatbib,final]{nips_2016}

\usepackage[utf8]{inputenc} 
\usepackage[T1]{fontenc}    
\usepackage{hyperref}       
\usepackage{url}            
\usepackage{booktabs}       
\usepackage{amsfonts}       
\usepackage{amsmath}

\title{Measuring Adverse Drug Effects on Multimorbity using Tractable Bayesian Networks }

\author{
Jessa Bekker\\
  KU Leuven\\
  Belgium\\
  \texttt{jessa.bekker@cs.kuleuven.be} \\
  \And
  Arjen Hommersom\\
  Open University of the Netherlands\\
  \& Radboud University Nijmegen\\
  Netherlands\\
  \texttt{arjenh@cs.ru.nl}\\
  \And{Martijn Lappenschaar}\\
  Radboud University Nijmegen\\
  Netherlands\\
  \texttt{mlappens@cs.ru.nl}\\
  \And
  Jesse Davis\\
  KU Leuven\\
  Belgium\\
  \texttt{jesse.davis@cs.kuleuven.be}\\
  }

\begin{document}

\maketitle

\begin{abstract}
Managing patients with multimorbidity often results in polypharmacy: the prescription of
multiple drugs. However, the long-term
effects of specific combinations of drugs and diseases are typically unknown. In particular, drugs prescribed for one
condition may result in adverse effects for the other. To investigate which types of drugs may affect the further progression 
of multimorbidity, we query models of diseases and prescriptions that are learned from primary
care data. State-of-the-art \emph{tractable} Bayesian network representations, on which such complex queries can be computed efficiently, are employed
for these large medical networks. 
Our results confirm that prescriptions may lead to unintended negative consequences in further development of multimorbidity in cardiovascular diseases.
Moreover, a drug treatment for one disease group may affect diseases of another group.
\end{abstract}

\section{Introduction}

There is currently an increased interest in
multimorbidity, i.e., the co-occurrence of at least two, but often more, chronic or acute diseases and medical conditions within a person~\cite{van1996comorbidity}. 
Due to the ageing population and other
socio-economic factors, its prevalence is increasing: over two-thirds of the elderly population in the Western world have at least three chronic conditions~\cite{barnett2012epidemiology}.
%

Clinicians who aim to manage patients with multiple diseases are confronted with
different guidelines, each 
meant to manage
an isolated disorder. 
By combining these recommendations, a patient may receive many different drugs. Unfortunately, it is not unlikely that a
drug prescribed for one condition may result in an adverse effect for the other~\cite{boyd-jama-2005,Boyd:2008ACER}.
In September 2016,  the
National Institute for Health and Care Excellence (NICE) in the United Kingdom
published the first clinical guideline on multimorbidity which emphasized patient-centred care~\cite{farmer2016clinical}.
Given the more generalist and
multidisciplinary nature of multimorbidity, 
clinical validation tools were recommended.

%


One significant issue with multimorbidity is polypharmacy~\cite{marengoni2015guidelines}.
Drugs are often introduced to prevent future morbidity and mortality, but in context of multimorbidity they may instead increase the burden.
Primary care e-health records can help identify markers of increased treatment burden,
but until now, there have been relatively few tools that indicate which drugs
may be harmful to a patient with multiple diseases.

We employ tractable Bayesian network learning to answer the main question in this paper:
what is the effect of various treatments on the further development of multimorbidity? 
To that end, we present the first method for learning Bayesian networks that can answer these complex type of questions.
Our learned network models interactions between multiple diseases and drug treatments in the primary care.
By identifying which treatment could potentially harm the patient and increase the burden of
multimorbidity even further, this work aimes to provide the first step towards
better recommendations to reduce the effects of treatment on the problem of multimorbidity.


\section{Methods}
In this paper, we investigate the effect of various treatments on further multimorbidity development. More concretely, we answer the question: Which drug treatment groups increase the probability of suffering from more diseases within a group, a group being either cardiovascular or musculo-skeletal diseases. Furthermore, we are interested in the number of additional diseases a patient typically gets.

\subsection{Data}

The data used for analysis were obtained from the Netherlands institute for health services
research Primary Care Database (NPCD, formerly known as the LINH database). 
All Dutch inhabitants are obligated to
register with a general practice, and the NPCD registry contains information
of routinely recorded data from about all patients of approximately 90 general
practices. Longitudinal data of approximately one and a half million patient
years, covering 2003-2011, were used in our analysis. 
Patient data is available for the whole time frame,
unless the patient moved out of the practice or the practice itself opted out.
Our analysis includes 222,506 patients, aged over 35 years. 166,881 of the patients were randomly picked as training data, 22,250 as validation data and 33,375 as test data. 

\begin{table}[ht]
\small
  \caption{Drug groups under consideration}
  \label{tab:drugGroups}
  \centering
  \begin{tabular}{llr}
    \toprule
    Code & Name     & Prevalence at baseline  \\
    \midrule
C01 &Cardiac therapy&	0.072	\\
C02 &Antihypertensives &	0.060	\\
C03 &Diuretics &	0.006	\\
C04 &Peripheral vasodilators&	0.142	\\
C05 &Vasoprotectives&	0.001	\\
C07 &Beta blocking agents&	0.037	\\
C08 &Calcium channel blockers&	0.162	\\
C09 &Agents acting on the renin-angiotensin system&	0.066	\\
C10 &Lipid modifying agents&	0.145	\\
M01 &  Anti-inflammatory and antirheumatic products &	0.123	\\
M02 &Topical products for joint and muscular pain&	0.333	\\
M03 & Muscle relaxants&	0.005	\\
M04 & Antigout preparations&	0.002	\\
M05 &Drugs for treatment of bone diseases&	0.011	\\
M09 &Other drugs for disorders of the musculo-skeletal system&	0.021	\\
N02 &Analgesics&	0.008	\\

    \bottomrule
  \end{tabular}
\end{table}

The data was split in two time-periods: $T_1 = [2003-2006]$ and $T_2 = [2007-2011]$.
We used $T_1$ as the situation at baseline, and $T_2$ as the situation we aim to predict.
We selected 16 cardiovascular and 26 musculo-skeletal diseases (respectively group K and L of the International Classification of Primary Care,
ICPC). Two groups were used so that the effects of 
treatments could be measured, both within the same group of diseases and between groups.
Furthermore, we considered 16 groups of drug treatment, listed in table~\ref{tab:drugGroups},
which are related to the two groups of diseases under consideration: all ATC codes
classified in the sections \emph{cardiovascular system} (C) and \emph{musculo-skeletal system} (M) . Additionaly, we included \emph{analgesics}
(N02), because we hypothesize that this drug
may be overprescribed for various conditions.

\subsection{Procedure}

We model the medical domain by a Bayesian network which we subsequently use for answering our research questions. Bayesian networks have been used before to investigate multimorbidity~\cite{Lappenschaar:AIME2013,Lappenschaar:NIVEL}. They are popular because of their compact and intuitive nature. However, using them for answering queries about the domain does not scale well, as this problem is in general NP-hard, even for simple marginal queries. Tractable learning is a new promising field that focuses on learning Bayesian networks and other types of probabilistic graphical models that guarantee to answer certain classes of queries within limited time.
Which classes of queries can be answered efficiently depends on the tractable representation. The most general representation that is being used today are Sentential Decision Diagrams (SDDs)~\cite{Darwiche:IJCAI2011,Bekker:NIPS15}. Its query space includes symmetric queries which aggregate over a subset of the variables. This allows us to reason on a higher level, for example, by asking queries about the class of cardiovascular diseases instead of a specific type of heart attack.


This paper presents the first method for learning Bayesian networks that efficiently answer symmetric queries efficiently, by using SDDs as their tractable representation. SDDs are highly flexible representations which facilitate adaptation of established non-tractable learners to tractable SDD learners. The Bayesian network is encoded as an SDD using ENC2 of Chavira and Darwiche (2008) \cite{Chavira:2008}. Ordering-based search, rather than structure search, was used, as it performs equally well and usually faster~\cite{Teyssier:UAI05}. Essentially, a fixed variable ordering facilitates learning the optimal network structure efficiently. Hence, searching for the best structure can be reduced to searching for the best ordering. The ordering, which is initially random, gets improved by greedily swapping variables that result in the best score. Local optima are avoided using restarts and Tabu lists. Tractable ordering-based search guarantees tractability by using a score that favors tractable models. Unfortunately, this extra score criterion makes it hard to learn the optimal network structure, even when the ordering is known. Nevertheless, searching over the order is advantageous because of its speed.

We adapted the algorithm to use decision trees for the conditional probability distributions (CPT-trees) instead of tables as they are shown to be able to capture complex domains with less variables and are therefore more tractable~\cite{Friedman:lgm}. To find a good structure for an ordering, the CPT-trees are greedily grown, optimizing the improvement in likelihood penalized with the decrease in tractability~\cite{lowd:AISTATS2013}. This continues until the likelihood converges or as long as learning time and tractability constraints allow it. Next, two variables are swapped in the ordering. The swap that provides the highest upperbound on likelihood increase, is selected. It is executed by removing all the edges to the swapped variables in the Bayesian network. This process of growing CPT-trees and swapping variables is repeated as long as time allows. Note that the last model is not necessarily the best one found by the process. A model is picked from all the seen models based on its validation set likelihood

We learned Bayesian networks for three days with 60 restarts. The parameters have a symmetric Dirichlet distribution with parameter $\alpha \in \{1.0,0.1,0.001,0.00001\}$ as a prior. The minimum contribution in log likelihood of a split needs to be 0.0001. The maximum SDD size is 2,000,000. The maximum time for adding splits is 900s. The tabu list size is 10.

The selected model is used to calculate the probability of suffering from additional diseases within a group, both in general and after receiving a specific drug treatment. These probabilities are compared to see which drug treatments increase the probability. To investigate the size of the increment, we calculate the probability of an increment of size $k \in \{1,2,3,4\}$. To calculate the probabilities, the SDD of the chosen model is queried, using the publically available software.\footnote{\url{https://dtai.cs.kuleuven.be/software/learnsdd/}}

\subsection{Analysis}

Table~\ref{tab:exp} gives the odds of suffering from additional diseases within a group, after receiving a certain drug treatment. It is calculated as follows:
\[
 \frac{
 \Pr(\# \text{diseases in group in }T_2 > \#\text{diseases in group in }T_1~|~\text{drug treatment in }T_1)
 }{
 \Pr(\#\text{diseases in group in }T_2 > \#\text{diseases in group in }T_1)
 }
\]

This number is 1 if the drug has no influence on the number of diseases and is greater or smaller than 1 when it respectively increases or decreases the number of diseases. The denominator of the fraction is 0.35 for cardiovascular and 0.44 for musculo-skeletal diseases. Big increases are marked in bold. Table~\ref{tab:expk} lists the probabilities of getting exactly $k$ more diseases when there is an increase.

\begin{table}[ht]

\small
\centering
 \caption{The odds of suffering from additional diseases within a group, after receiving a certain drug treatment.}
 \label{tab:exp}
 \begin{tabular}{|c|cc||c|cc|}
 \hline
Drug group	&	\multicolumn{2}{c||}{Disease group}	& Drug group	&	\multicolumn{2}{c|}{Disease group} \\
	&	Cardiovascular	&	Musculo-skeletal	&		&	Cardiovascular	&	Musculo-skeletal	\\\hline
C01	&	\bf1.17	&	0.89	&	M01	&	1.02	&	0.92	\\
C02	&	0.98	&	0.95	&	M02	&	\bf1.20	&	0.76	\\
C03	&	\bf1.18	&	1.01	&	M03	&	1.00	&	1.00	\\
C04	&	0.89	&	0.88	&	M04	&	1.00	&	1.00	\\
C05	&	1.00	&	1.00	&	M05	&	0.95	&	0.95	\\
C07	&	1.00	&	0.93	&	M09	&	\bf1.30	&	0.85	\\
C08	&	0.98	&	0.94	&		&		&		\\
C09	&	0.89	&	0.91	&	N02	&	\bf1.46	&	0.88	\\
C10	&	0.89	&	0.90	&		&		&		\\

\hline
 \end{tabular}

\end{table}

\begin{table}[ht]

\small
 \centering
 \caption{The probabilities of getting exactly $k$ more diseases when there is an increase.}
 \label{tab:expk}
 \begin{tabular}{|c|cccc|}
  \hline
  k	&	1	&	2	&	3	&	4\\\hline
  $\Pr(k$ more cardiovascular diseases$)$ 	& 0.68	& 0.23	& 0.07	& 0.02 \\
  $\Pr(k$ more musculo-skeletal diseases$)$	& 0.47	& 0.28	& 0.14	& 0.07\\
\hline
 \end{tabular}

\end{table}


\section{Results}

Prescriptions may lead to unintended negative consequences in further development of multimorbidity. Our results indeed show that there is a relationship between drugs and increase of cardiovascular diseases, but not of musculo-skeletal diseases. Notably, prescriptions may affect diseases in different subgroup from the disease that is being treated. For example, drugs for disorders of the musculo-skeletal system (M02 and M09) and analgesics (N02) seem to have a negative effect on cardiovascular health. These results are not sufficient to conclude causal relationships but they do emphasize the importance of a multidisciplinary approach in the context of multimorbidity. Note that these negative effects may be different between subgroups of diseases. Our results in table~\ref{tab:expk} suggest that if cardiovascular health decreases, the number of additional diagnoses is limited compared to the musculo-skeletal subgroup. 

\section{Conclusion}

In this paper we investigated the effect of various treatments on further multimorbidity development, employing Bayesian network learning to model the domain. Although Bayesian networks have been used before to investigate multimorbidity,
scaling such an analysis to larger networks which include prescriptions, requires tractable representations~\cite{Lappenschaar:AIME2013,Lappenschaar:NIVEL}.
In this work we examined two groups of diseases and discovered that prescription for one group can negatively effect the multimorbidity of the other group. Furthermore, these effects may differ between groups.
This underlines the importance of a multidisciplinary approach in the context of multimorbidity. Future work will include analysis of more types of diseases and drug treatments.

\section*{Acknowledgements}
 We thank Guy Van den Broeck for helpful suggestions. Jessa Bekker is supported by IWT (SB/141744). Jesse Davis is partially supported by the Research Fund KU Leuven (OT/11/051, C22/15/015), EU FP7 Marie Curie CIG (\#294068), IWT (SBO-HYMOP) and FWO-Vlaanderen (G.0356.12).

\small
\bibliography{paper} 
\bibliographystyle{plain}

\end{document}